\documentclass[sigconf]{acmart}
\AtBeginDocument{%
  }

\copyrightyear{2025}
\acmYear{2025}
\setcopyright{acmlicensed}\acmConference[CIKM '25]{Proceedings of the 34th ACM International Conference on Information and Knowledge Management}{November 10--14, 2025}{Seoul, Republic of Korea}
\acmBooktitle{Proceedings of the 34th ACM International Conference on Information and Knowledge Management (CIKM '25), November 10--14, 2025, Seoul, Republic of Korea}
\acmDOI{10.1145/3746252.3761476}
\acmISBN{979-8-4007-2040-6/2025/11}

\usepackage{hyphenat}
\usepackage{hyperref}
\usepackage{listings}
\usepackage{xcolor}
\usepackage{caption}
\usepackage{booktabs}

\definecolor{codeblue}{rgb}{0.0, 0.0, 0.6}
\definecolor{codegreen}{rgb}{0.0, 0.5, 0.0}
\definecolor{codegray}{rgb}{0.4, 0.4, 0.4}
\definecolor{codepurple}{rgb}{0.58, 0.0, 0.82}
\definecolor{backcolor}{rgb}{0.98, 0.98, 0.99}
\definecolor{importpink}{rgb}{0.8, 0.2, 0.6}
\definecolor{constraintred}{rgb}{0.58,0,0.82}

\lstdefinestyle{mystyle}{
    language=Python,
    backgroundcolor=\color{backcolor},   
    keywordstyle=\color{codeblue}\bfseries,
    commentstyle=\color{codegreen},
    stringstyle=\color{constraintred},
    numberstyle=\tiny\color{codegray},
    basicstyle=\ttfamily\small,
    numbers=left,
    stepnumber=1,
    numbersep=-10pt,
    breaklines=true,
    breakatwhitespace=true,
    showstringspaces=false,
    frame=tb,
    framerule=0.5pt,
    framexleftmargin=0.2em,
    xleftmargin=0.1em,
    xrightmargin=0.5em,
    columns=fullflexible,
    linewidth=\columnwidth
}

\DeclareCaptionFormat{nocaption}{#1#2#3\par}
\usepackage[colorinlistoftodos]{todonotes}

\usepackage{tikz}
\newcommand{\circlednum}[1]{%
  \tikz[baseline=(char.base)]{
    \node[shape=circle,draw,inner sep=0.8pt] (char) {\footnotesize #1};}}

\begin{document}

\title{MMM-fair: An Interactive Toolkit for Exploring and Operationalizing Multi-Fairness Trade-offs}

\author{Swati Swati}
\authornote{Both authors contributed equally to this research.}
\email{swati.swati@unibw.de}
\orcid{0000-0002-7637-6640}
\affiliation{%
  \institution{University of the Bundeswehr}
  \city{Munich}
  \country{Germany}
}

\author{Arjun Roy}
\authornotemark[1]
\email{arjun.roy@unibw.de}
\orcid{0000-0002-4279-9442}
\affiliation{%
  \institution{University of the Bundeswehr}
  \city{Munich}
  \country{Germany}
}
\affiliation{%
  \institution{Free University Berlin}
  \city{Berlin}
  \country{Germany}
}

\author{Emmanouil Panagiotou}
\email{emmanouil.panagiotou@fu-berlin.de}
\orcid{0000-0001-9134-9387}
\affiliation{%
  \institution{Free University Berlin}
  \city{Berlin}
  \country{Germany}
}
\affiliation{%
  \institution{University of the Bundeswehr}
  \city{Munich}
  \country{Germany}
}

\author{Eirini Ntoutsi}
\email{eirini.ntoutsi@unibw.de}
\orcid{0000-0001-5729-1003}
\affiliation{%
  \institution{University of the Bundeswehr}
  \city{Munich}
  \country{Germany}
}

\renewcommand{\shortauthors}{Swati et al.}


\begin{abstract}
    Fairness-aware classification requires balancing performance and fairness, often intensified by intersectional biases. Conflicting fairness definitions further complicate the task, making it difficult to identify universally fair solutions. Despite growing regulatory and societal demands for equitable AI, popular toolkits offer limited support for exploring multi-dimensional fairness and related trade-offs. To address this, we present \emph{mmm-fair}, an open-source toolkit leveraging boosting-based ensemble approaches that dynamically optimizes model weights to jointly minimize classification errors and diverse fairness violations, enabling flexible multi-objective optimization. The system empowers users to deploy models that align with their context-specific needs while reliably uncovering intersectional biases often missed by state-of-the-art methods. In a nutshell, mmm-fair uniquely combines in-depth multi-attribute fairness, multi-objective optimization, a no-code, chat-based interface, LLM-powered explanations, interactive Pareto exploration for model selection, custom fairness constraint definition, and deployment-ready models in a single open-source toolkit, a combination rarely found in existing fairness tools. Demo walkthrough available at: \href{https://youtu.be/_rcpjlXFqkw}{\textcolor{blue}{https://youtu.be/\_rcpjlXFqkw}}.

\end{abstract}




\maketitle

\section{Introduction}

    \begin{table*}[htbp]
        \centering
        \caption{Feature-based comparison of \emph{mmm-fair} and related popular fairness-aware toolkits.}
        \label{tab:feature-grid}
        \resizebox{\textwidth}{!}{
        \setlength{\tabcolsep}{3pt} 
        {\small 
        \begin{tabular}{l c c c c c c c c c}
            \toprule
            
            \textbf{Feature} & \textbf{mmm-fair} & \textbf{IBM AIF360} & \textbf{MS Fairlearn} & \textbf{Google What-If} & \textbf{Snowflake-TruEra} 
            & \textbf{WhyLabs} & \textbf{Fiddler} & \textbf{FairBench} \\

             & (Ours) & \cite{bellamy2019ai} & \cite{bird2020fairlearn} & \cite{wexler2019if} & \cite{truera_ai_quality} &
             \cite{whylabs2025} & \cite{gade2020explainable} &\cite{krasanakis2024standardizing} \\

            \midrule
            In-depth multi-attribute fairness & \textcolor{green!60!black}{\textbf{$\checkmark$}} & \textcolor{green!60!black}{\textbf{$\checkmark$}} & \textcolor{green!60!black}{\textbf{$\checkmark$}} & \textcolor{gray}{\textbf{$\circ$}} & \textcolor{gray}{\textbf{$\circ$}} 
            & \textcolor{gray}{\textbf{$\circ$}} & \textcolor{gray}{\textbf{$\circ$}} & \textcolor{gray}{\textbf{$\circ$}}\\
            Multi-objective optimization & \textcolor{green!60!black}{\textbf{$\checkmark$}} & \textcolor{red}{--} & \textcolor{red}{--} & \textcolor{gray}{\textbf{$\circ$}} & \textcolor{red}{--} 
            & \textcolor{red}{--} & \textcolor{red}{--} & \textcolor{gray}{\textbf{$\circ$}}\\
            No-code user interface & \textcolor{green!60!black}{\textbf{$\checkmark$}} & \textcolor{red}{--} & \textcolor{green!60!black}{\textbf{$\checkmark$}} & \textcolor{green!60!black}{\textbf{$\checkmark$}} & \textcolor{green!60!black}{\textbf{$\checkmark$}} 
            & \textcolor{green!60!black}{\textbf{$\checkmark$}} & \textcolor{red}{--} & \textcolor{green!60!black}{\textbf{$\checkmark$}}\\
            Chat-based interaction & \textcolor{green!60!black}{\textbf{$\checkmark$}} & \textcolor{red}{--} & \textcolor{red}{--} & \textcolor{red}{--} & \textcolor{red}{--} 
            & \textcolor{green!60!black}{\textbf{$\checkmark$}} & \textcolor{gray}{\textbf{$\circ$}} & \textcolor{red}{--}\\
            Pareto trade-off explorer & \textcolor{green!60!black}{\textbf{$\checkmark$}} & \textcolor{red}{--} & \textcolor{red}{--} & \textcolor{orange}{\textbf{$\triangle$}} & \textcolor{red}{--} 
            & \textcolor{red}{--} & \textcolor{red}{--} & \textcolor{orange}{\textbf{$\triangle$}}\\
            Custom constraints (DP, EO, etc.) & \textcolor{green!60!black}{\textbf{$\checkmark$}} & \textcolor{green!60!black}{\textbf{$\checkmark$}} & \textcolor{green!60!black}{\textbf{$\checkmark$}} & \textcolor{gray}{\textbf{$\circ$}} & \textcolor{gray}{\textbf{$\circ$}} 
            & \textcolor{gray}{\textbf{$\circ$}} & \textcolor{gray}{\textbf{$\circ$}} & \textcolor{gray}{\textbf{$\circ$}}\\
            Fairness-aware training, not just auditing & \textcolor{green!60!black}{\textbf{$\checkmark$}} & \textcolor{green!60!black}{\textbf{$\checkmark$}} & \textcolor{green!60!black}{\textbf{$\checkmark$}} & \textcolor{red}{--} & \textcolor{red}{--} 
            & \textcolor{red}{--} & \textcolor{red}{--} & \textcolor{red}{--}\\
            Deployment-ready models & \textcolor{green!60!black}{\textbf{$\checkmark$}} & \textcolor{red}{--} & \textcolor{red}{--} & \textcolor{red}{--} & \textcolor{red}{--} 
            & \textcolor{red}{--} & \textcolor{red}{--} & \textcolor{red}{--}\\
            Open source & \textcolor{green!60!black}{\textbf{$\checkmark$}} & \textcolor{green!60!black}{\textbf{$\checkmark$}} & \textcolor{green!60!black}{\textbf{$\checkmark$}} & \textcolor{green!60!black}{\textbf{$\checkmark$}} & \textcolor{red}{--} 
            & \textcolor{orange}{\textbf{$\triangle$}} & \textcolor{orange}{\textbf{$\triangle$}} & \textcolor{green!60!black}{\textbf{$\checkmark$}}\\
            LLM-based explanations & \textcolor{green!60!black}{\textbf{$\checkmark$}} & \textcolor{red}{--} & \textcolor{red}{--} & \textcolor{red}{--} & \textcolor{green!60!black}{\textbf{$\checkmark$}} 
            & \textcolor{green!60!black}{\textbf{$\checkmark$}} & \textcolor{orange}{\textbf{$\triangle$}} & \textcolor{red}{--}\\
            \bottomrule
        \end{tabular}
        }
        }
        \vspace{0.5ex} 
        {\footnotesize %
        \textbf{Note:} \textcolor{green!60!black}{\textbf{$\checkmark$}} Fully Supported; \textcolor{orange}{\textbf{$\triangle$}} Partial/Basic Support; \textcolor{gray}{\textbf{$\circ$}} Reporting Only; \textcolor{red}{--} Not Supported.\par}
    \end{table*}
    
    \textbf{Motivation.} As machine learning (ML) systems are increasingly deployed across critical domains such as healthcare, finance, and hiring, concerns about biased or unfair outcomes have intensified~\cite{bellamy2019ai,swati2024ourmultimodal,chen2023recruitment}. These concerns arise from the growing recognition that algorithmic systems can inadvertently reproduce existing societal inequalities or introduce new forms of discrimination, often disproportionately affecting historically marginalized groups~\cite{barocas_fairmlbook,RoyFact23}. Fairness-aware classification presents an inherently complex, multi-objective challenge: it requires balancing predictive performance with fairness constraints across multiple protected attributes, often under conditions of class or group imbalance~\cite{le2022survey}. This challenge is further compounded by multiple, often incompatible definitions of fairness~\cite{10.1145/3433949}, making it difficult to identify solutions that are both universally acceptable and practically effective.

    \noindent\textbf{Solution overview.} To address the multifaceted challenges of fairness-aware classification, we introduce \emph{mmm-fair}, a Python package designed to support flexible, interpretable, and context-sensitive model development. It provides comprehensive support for \emph{multi-fairness} (the simultaneous consideration of \emph{multiple protected attributes}, \emph{fairness definitions}, and \emph{optimization objectives}) within a unified framework. The system features an interactive, modular workflow that operationalizes fairness as a practical, end-to-end process, guiding users through data profiling, subgroup analysis, model training, trade-off exploration, LLM-based explanation, and deployment. Among its key components is the integrated Pareto front explorer, which enables users to evaluate trade-offs across multiple fairness and performance metrics and select models aligned with user-defined goals. A chat-based interface further facilitates explanation and supports iterative refinement throughout the workflow. 
    \emph{Why mmm-fair? Because “What’s fair?” should be the user’s choice, not the algorithm’s.} 

    \noindent\textbf{Comparison with existing works.} Although existing fairness toolkits have contributed significantly to the field, they often fall short in their ability to provide comprehensive multi-fairness support within a single, integrated framework. As shown in Table~\ref{tab:feature-grid}, \emph{mmm-fair} stands out from popular toolkits by offering a broader and more versatile set of capabilities. Empirical results on benchmark datasets further demonstrate that \emph{mmm-fair} reliably uncovers intersectional biases and reduces group disparities without sacrificing accuracy or increasing overfitting. It maintains robust performance even in imbalanced and complex real-world datasets, establishing itself as a uniquely effective and versatile solution.
    
    \noindent\textbf{System Availability.} \emph{mmm-fair} is publicly available: as a package on \textbf{PyPI} (\href{https://pypi.org/project/mmm-fair/}{\textcolor{blue}{https://pypi.org/project/mmm-fair/}}), with source code on \textbf{GitHub} (\href{https://github.com/arjunroyihrpa/MMM_fair}{\textcolor{blue}{https://github.com/arjunroyihrpa/MMM\_fair}}), and a demonstration video at \textbf{YouTube} (\href{https://youtu.be/_rcpjlXFqkw}{\textcolor{blue}{https://youtu.be/\_rcpjlXFqkw}}).

\section{mmm-fair: Package Overview}
    
    Built on the foundational work of~\citet{roy2022mmm}, \emph{mmm-fair} is a Python toolkit that generalizes the “Multi-fairness Under Class-Imbalance” approach to support a broader set of fairness definitions, optimization objectives, and experimental configurations within a unified framework. While the original work focused on Disparate Mistreatment, it extends its support to widely used fairness criteria such as Demographic Parity, Equalized Odds, and others, offering enhanced flexibility and customization during model training. At its core is a boosting-based ensemble method that adjusts model weights to jointly minimize classification error and fairness violations. This architecture enables efficient exploration of trade-offs across multi-fairness constraints and incorporates the following key functionalities:

    \begin{itemize}
        \item \textbf{Multi-attribute fairness:} allows fairness assessment across multiple protected attributes, such as age, race, and gender.
        \item \textbf{Multiple fairness definitions:} enables selection among demographic parity, equal opportunity, equalized odds, and others.
        \item \textbf{Fairness-integrated boosting:} leverages AdaBoost-style and gradient-boosted ensembles with a fairness-weighted objective, controlled by the hyper-parameter \emph{gamma}, to jointly optimize predictive accuracy and fairness during training.
        \item \textbf{Pareto front-based model selection:} offers identification and visualization of optimal trade-offs between fairness and predictive performance across multiple objectives, fairness definitions, and protected attributes.
        \item \textbf{Adaptive handling of difficult instances:} dynamically adjusts emphasis on difficult samples to reduce over-correction and improve stability, particularly once fairness goals are partially met.
        \item \textbf{Extensible modular design:} allows easy integration of fairness metrics, definitions, and base learners through scikit-learn APIs and CLI/code, ensuring adaptability and maintainability.
        \item \textbf{Chat-based user interface:} provides a simple, no-code interface for interacting with the toolkit and receiving explanations and guidance.
        \item \textbf{Seamless model deployment:} enables export of trained models for downstream use and application integration.
        \item \textbf{User-friendly and open source:} offers intuitive interface, step-by-step guidance, and publicly available reproducible workflows.
    \end{itemize}

    Previous fairness-aware boosting approaches, such as fairness-aware AdaBoost variants~\cite{iosifidis2022parity} and MFBPP~\cite{roy2022mmm}, have typically focused on a single fairness definition and relied primarily on reweighting strategies. In contrast, our package offers greater flexibility by allowing users to choose from multiple fairness definitions and directly incorporating the selected constraint across multiple protected attributes into the gradient boosting objective. To achieve joint optimization, we introduce a softmax-weighted aggregation of fairness gradients across attributes, enabling the model to balance predictive performance with fairness requirements across multiple attributes. Finally, we adopt a Pareto-front model selection strategy that explicitly balances accuracy, class imbalance, and fairness in a principled manner, moving beyond ad-hoc or purely post-hoc adjustments.
    
    This multifaceted architecture enables mmm-fair to support a broad spectrum of real-world applications where fairness is essential. By decoupling the boosting strategy from any single definition, the toolkit allows users to flexibly explore and prioritize fairness, performance, or a tailored balance of both.  

\section{Using mmm-fair}
    
    \emph{mmm-fair} is a comprehensive toolkit for fairness-aware machine learning, designed to address real-world challenges across diverse application domains, with an interactive user interface and command-line options to suit varied user needs.

    \vspace{0.5em}
    \noindent \textbf{Application domains:} \emph{mmm-fair} applies to a wide range of domains, including credit scoring, healthcare, hiring, and any scenario where fairness across multiple protected groups and definitions is required~\cite{le2022survey}. Typical use cases include evaluating fairness-accuracy trade-offs in imbalanced datasets, mitigating algorithmic bias for regulatory compliance, and analyzing subgroup outcomes in high-stakes decision-making systems.
    
    \vspace{0.5em}
    \noindent \textbf{Installation.} The toolkit offers both an intuitive web-based interface and a command-line interface (CLI), making it accessible to users with diverse technical backgrounds. To install, use: \colorbox{blue!5}{\textcolor{black}{\texttt{pip install mmm-fair}}}. Once installed, the CLI is particularly suited for machine learning practitioners who prefer scripting and need precise control over model configurations and fairness constraints. The following script demonstrates a minimal working example for setting fairness objectives and training a model:

        \begin{lstlisting}[style=mystyle, caption={Quickstart: Defining Fairness Constraints and Training.}, label={lst:train}]
            from mmm_fair import MMM_Fair
            from sklearn.tree import DecisionTreeClassifier
            
            mmm = MMM_Fair(
            estimator=DecisionTreeClassifier(max_depth=5),
                constraints="EO", # or "DP", "EP", etc. 
                saIndex=sa_index, # (n_samples, n_protected)
                saValue=sa_value # dictionary or None
                # other parameters, e.g. alpha, gamma, etc.
            )
            
            mmm.fit(X, y) # X: features, y: labels
            y_pred = mmm.predict(X_test)
        \end{lstlisting}

        \begin{figure*}[t]
        \centering
        \includegraphics[width=\textwidth]{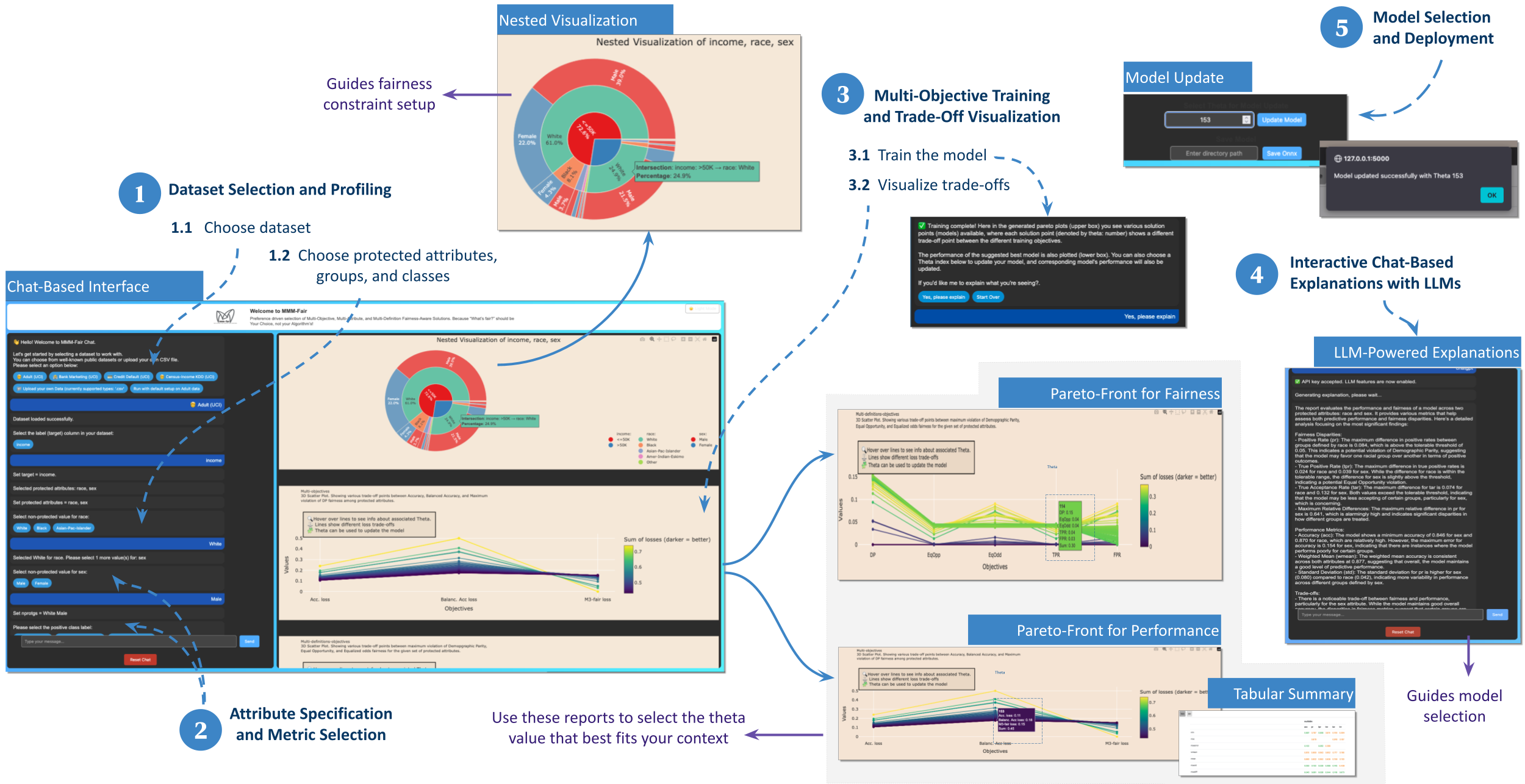}
        \caption{End-to-end pipeline for exploring and operationalizing multi-fairness trade-offs in \textit{mmm-fair}.}
        \label{fig:mmmfair-overview}
    \end{figure*}

    \subsection{An Interactive Demonstration}
        
        \noindent\textbf{A Real-World Scenario as an Example.}  
        Consider a cross\hyp{}functional team of data scientists, policy analysts, compliance experts, software engineers, and other stakeholders at a financial institution. Their goal is to develop a system that maintains high predictive performance while satisfying fairness constraints and policy requirements. The team must ensure that the model behaves equitably across protected attributes and their intersections, such as age, income, and education, using multiple fairness definitions. These challenges intensify with imbalanced data, overlapping subgroups, and competing objectives. \emph{mmm-fair} supports such teams in operationalizing multi-dimensional fairness through a seamless, interactive workflow. The following steps illustrate how mmm-fair can be used to systematically address these complex challenges in practice.
    
        \vspace{0.5em}
        \noindent \textbf{\circlednum{1} Dataset Selection and Profiling.} The workflow begins with users selecting a built-in benchmark, such as Adult Income or German Credit, or uploading a custom dataset. Once loaded, it generates an interactive, nested visualization that displays the distribution of key protected attributes and their intersections with the target variable; users can explore these profiles to immediately identify how different subgroups are represented, revealing imbalances or underrepresented groups that may raise fairness concerns. This initial exploration provides a clearer understanding of the dataset’s protected structure before moving on to fairness configuration and model development.
        
        \vspace{0.5em}
        \noindent \textbf{\circlednum{2} Attribute Specification and Metric Selection.} This step empowers users to define protected attributes and subgroups relevant to their context. The interface enables selection of a diverse set of features, such as gender, age, race, or education, to capture nuanced sources of bias. Users can also define fairness and performance metrics to guide evaluation, choosing from a diverse set of options like demographic parity, equalized odds, and balanced accuracy, among others. Metrics are prioritized as constraints, ensuring alignment with the goals of the cross-functional team.
        
        \vspace{0.5em}
        \noindent \textbf{\circlednum{3} Multi-Objective Training and Trade-Off Visualization.} Once objectives are set, the workflow advances to model training, producing a diverse set of candidate models in a single run. \emph{mmm-fair} then generates the Pareto front, clearly visualizing trade-offs between fairness and accuracy (see Figure~\ref{fig:mmmfair-overview}). The interface enables users to inspect metric values, compare alternatives, and immediately see how different configurations affect competing objectives. As users explore these plots, they often discover solutions or trade-offs not anticipated during setup. This interactive process supports transparent, interpretable, and robust model selection, ensuring alignment with institutional policies and stakeholder values.
    
        \vspace{0.5em}
        \noindent \textbf{\circlednum{4} Interactive Chat-Based Explanations with LLMs.} After model training and visualization, users can request explanations through the integrated chat interface; the system prompts for a provider, such as OpenAI or ChatGPT, and generates clear, context-aware natural language summaries. Users may ask follow-up questions, making it easy to interpret results and communicate with both technical and non-technical audiences.

        Importantly, the LLM used here does not compute statistical outputs; instead, it functions solely as a narrative layer that renders structured results into natural language. To minimize the potential for hallucinations, the module combines constrained prompting, structured response formats, and an algorithm-first pipeline with precomputed summaries. Explanations are shown alongside raw plots and metrics, allowing users to verify results directly. Together, these measures strengthen reliability and transparency while maintaining usability.
        
        \vspace{0.5em}
        \noindent \textbf{\circlednum{5} Model Selection and Deployment.} Once users identify the most suitable solution, usually by selecting a trade-off value (theta) along the Pareto front, they can save the model directly from the interface. This ensures the final configuration matches the defined fairness and performance priorities. Users can also export all relevant plots and charts for documentation and future audits.

        \noindent In this credit scoring scenario, \emph{mmm-fair} enables data scientists to systematically explore model trade-offs and identify solutions that best balance accuracy and fairness across different demographic groups. Policy analysts can assess how various fairness definitions apply to intersecting attributes, ensuring that institutional priorities and regulatory guidelines are reflected in the final model. Compliance experts benefit from transparent reporting and intuitive explanations, which support audit readiness and strengthen accountability. Non-technical stakeholders, including managers and external partners, can understand the impact of each modeling decision through clear visualizations and natural language summaries.  
    
        This collaborative and interpretable approach is not limited to credit scoring. Demonstrated through a realistic end-to-end scenario, \emph{mmm-fair}'s unified workflow and rich feature set guide users from dataset selection to model deployment. By combining multi-attribute fairness analysis, multi-objective optimization, and interactive exploration in a single interface, the toolkit empowers users to surface hidden biases, understand fairness-performance trade-offs, and select models that reflect their institutional goals. Rather than offering yet another fairness metric calculator, \emph{mmm-fair} equips practitioners with a hands-on, decision-support system for building fair, auditable, and deployment-ready models. It turns fairness from an abstract ideal into a concrete, actionable outcome. Preliminary internal evaluations indicate that the workflow facilitates adoption across diverse user groups, with usability studies planned to assess its effectiveness.


\section{Conclusion and Future Work}

    In this work, we introduce \emph{mmm-fair}, an interactive Python toolkit for exploring and operationalizing multi-fairness trade-offs in fairness-aware machine learning. It combines a boosting-based ensemble approach with a no-code, chat-driven interface, integrated LLM explanations, interactive Pareto front visualizations, and deployment-ready models, enabling users to define fairness constraints, examine subgroup outcomes, and address intersectional bias while preserving predictive accuracy.

    \noindent\textbf{Limitations and Future Work.} The current implementation supports only tabular data with manually specified protected attributes; the explanation module relies on external LLMs with user-provided API keys. At present, it is best suited for small to medium-scale analyses and does not yet support causal inference, longitudinal modeling, or large-scale deployment. Planned extensions include desktop and cloud-based interfaces, support for local models and hybrid explanation methods, and broader applicability to vision, text, and other non-tabular modalities.
    
\section*{Ethics statement.} 
    \emph{mmm-fair} promotes transparency and aims to reduce algorithmic bias; however, users must choose fairness definitions and interpret results with care, as societal contexts vary. While it does not guarantee universal fairness, the toolkit empowers practitioners to explore trade-offs and make informed decisions. 
\begin{acks}
    This research work is funded by the European Union under the Horizon Europe MAMMOth project, Grant Agreement ID: 101070285. UK participant in Horizon Europe Project MAMMOth is supported by UKRI grant number 10041914 (Trilateral Research LTD). The research is also supported by the EU Horizon Europe project STELAR, Grant Agreement ID: 101070122.
\end{acks}

\section*{GenAI Usage Disclosure}
Generative AI tools (e.g., ChatGPT and GPT-4) were employed to assist with the preparation of this manuscript. Specifically, these tools were utilized for text refinement, editing for clarity and readability, and generating illustrative examples. No original intellectual contributions or novel results were generated by these tools. The authors confirm that all AI-generated content has been reviewed, verified for accuracy, and integrated thoughtfully into the manuscript, ensuring full compliance with ACM’s Policy on Authorship, available at:
\url{https://www.acm.org/publications/policies/new-acm-policy-on-authorship}.

\bibliographystyle{ACM-Reference-Format}
\bibliography{mybibliography}
\end{document}